\documentclass[]{article} 
\usepackage{graphicx}
\usepackage{amssymb}
\usepackage{amsmath}

\usepackage{stmaryrd}
\usepackage{wasysym}
\usepackage{extarrows}

\usepackage{proof}
\usepackage{lambek-fs}

\renewcommand{\oslash}{\mathbin{\scalebox{.85}{$\varoslash$}}}
\renewcommand{\obslash}{\mathbin{\scalebox{.85}{$\varobslash$}}}

\newcommand{\W}[1]{\textsf{#1}}

\newcommand{\nd}[2]{#1 \vdash #2}

\newcommand{\seq}[2]{#1\Rightarrow #2}

\newcommand{\arr}[2]{#1\vdash #2}

\newcommand{\Ra}{\!\shortrightarrow\!}

\newcommand{\bs}{\backslash}
\newcommand{\comu}{\widetilde{\mu}}

\newcommand{\CBV}[1]{\lceil #1\rceil}

\newcommand{\CBVlex}[1]{\llceil #1\rrceil}

\newcommand{\cut}[1]{}

\newcommand{\LG}{\textbf{LG}}
\renewcommand{\arr}[2]{#1 \rightarrow #2}

\newcommand{\fdia}{\Diamond}
\newcommand{\gbox}{\Box}

\newcommand{\LaMu}{\mbox{$\overline{\lambda}\mu\comu$}}

 \newcommand{\leftleftharpoons}{%
  \mathbin{\ooalign{\hfil\raisebox{1pt}{$\leftharpoonup$}\hfil\cr\hfil
  \raisebox{-1pt}{$\leftharpoondown$}\hfil\crcr}}} 

 \newcommand{\rightrightharpoons}{%
  \mathbin{\ooalign{\hfil\raisebox{1pt}{$\rightharpoonup$}\hfil\cr\hfil
  \raisebox{-1pt}{$\rightharpoondown$}\hfil\crcr}}}

\newcommand{\otimesS}{\cdot\otimes\cdot}
\newcommand{\slashS}{\cdot\slash\cdot}
\newcommand{\bsS}{\cdot\bs\cdot}
\newcommand{\oplusS}{\cdot\oplus\cdot}
\newcommand{\oslashS}{\cdot\oslash\cdot}
\newcommand{\obslashS}{\cdot\obslash\cdot}

\newcommand{\lneg}[1]{{}^{\mathbf{0}}#1}
\newcommand{\rneg}[1]{#1^{\mathbf{0}}}
\newcommand{\ldneg}[1]{{}^{\mathbf{1}}#1}
\newcommand{\rdneg}[1]{#1^{\mathbf{1}}}

\title{Symmetric categorial grammar:\\[1ex]
	residuation and Galois connections}
\author{Michael Moortgat}
\institution{Utrecht Institute of Linguistics OTS}
\begin{document}
\maketitle

\section{Introduction}\label{intro}
In a paper written in 1983, V.N.~Grishin proposed
to complement the product, left and right division operations of Lambek's syntactic calculus
with a dual set of operations: coproduct, and the subtraction operations of right and left difference.
In its most elementary form, the resulting categorial type logic, which we'll
refer to as the Lambek-Grishin calculus (\LG), is given by the preorder axioms for
the derivability arrow $\rightarrow$, together with the invertible rules of inference
below, characterizing the operations $\otimes,\slash,\backslash$ as a residuated triple,
and $\oplus,\obslash,\oslash$ as a dual residuated triple.
\[\begin{array}{r@{\quad\Leftrightarrow\quad}c@{\quad\Leftrightarrow\quad}l}
\arr{A}{C/B} & \arr{A\otimes B}{C} & \arr{B}{A\bs C}\\
\arr{B\obslash C}{A} & \arr{C}{B\oplus A} & \arr{C\oslash A}{B}\\
\end{array}\]
From this basis, extended versions can be obtained in terms of
linear \emph{distributivity principles}. These allow for interaction between
the $\otimes$ and $\oplus$ families while preserving their individual (non-commutative,
non-associative) characteristics.

\LG\ exhibits two kinds of symmetry, given by the translation tables below.\footnote{Abbreviating a
long list of definitional equations $(C\slash D)^{\bowtie}=D^{\bowtie}\backslash C^{\bowtie}$,
$(D\backslash C)^{\bowtie}=C^{\bowtie}\slash D^{\bowtie}$, \ldots\ For atoms, $p^{\bowtie}=p=p^{\infty}$.}
We write ${\cdot}^{\bowtie}$ for the left-right symmetry of the
original syntactic calculus; it preserves derivability: $\arr{A}{B}$ iff $\arr{A^{\bowtie}}{B^{\bowtie}}$.
The ${\cdot}^{\infty}$ symmetry relates the operations of the $\otimes$ family to their
duals. This symmetry is arrow-reversing: $\arr{A}{B}$ iff $\arr{B^{\infty}}{A^{\infty}}$.
\[
\begin{array}{cc}
\bowtie & \begin{array}{c@{\quad}c@{\quad}c@{\quad}c@{\quad}c@{\quad}c}
	C/D & A\otimes B & B\oplus A & D\obslash C\\ \hline\hline
	D\bs C & B\otimes A & A\oplus B & C\oslash D\\
	\end{array}\\
\end{array}\]
\[\begin{array}{cc}
\infty & \begin{array}{c@{\quad}c@{\quad}c}
	C/B & A\otimes B & A\bs C \\ \hline\hline
	B\obslash C & B\oplus A & C\oslash A \\
	\end{array}\\
\end{array}
\]

With his 1993 paper, Jim Lambek was among the first to bring Grishin's work to the
attention of a wider audience; also he had the paper translated by his
student \v{C}ubri\'c so as to make it accessible for researchers lacking
fluency in Russian.\footnote{The translation introduces little puzzles
of its own. In the references, the author of a well-known study on partially
ordered algebraic systems appears in disguise as L.~Fooks --- the English
transliteration of the Russian transliteration doesn't quite
disclose the identity of the Hungarian mathematician with the German name.}
Linguistic exploration is of a more recent date. In \cite{jfak60}, I give
a survey of results obtained so far. \emph{Semantically}, \LG\ derivations
are associated with terms of the linear lambda calculus, as is the case
for the original categorial grammars. But because \LG\ logically
is a multiple-conclusion system, the target terms 
are obtained via a continuation-passing-style translation into multiplicative
intuitionistic linear logic. The translation introduces a
distinction between \emph{values} and \emph{contexts} of evaluation; the context
is explicitly included into the meaning composition process. Recent work in
formal semantics (e.g.~\citep{degr:type01,barker2006types}) has forcefully argued for
this view on the syntax-semantics
interface. Symmetric \LG\ provides a solid prooftheoretic basis for a
continuation semantics, and for the different evaluation strategies
that go with it. \emph{Syntactically}, Grishin's distributivity principles
make it possible to interleave the composition of phrases out of their
constituent parts with the composition of evaluation contexts for the
semantic values associated with these phrases. This creates new
possibilities for handling \emph{discontinuous} dependencies that
arise when syntactic and semantic composition are out of tune.

My aim in this paper is to complement the symmetry between (dual) residuated
type-forming operations with an orthogonal opposition that
contrasts residuated and \emph{Galois connected} operations. Whereas the (dual)
residuated operations are monotone, the Galois connected
operations (and their duals) are antitone. The paper is organized
as follows. In \S\ref{resgal}, the vocabulary is extended
with a Galois connected pair and a dual Galois connected pair,
and the algebraic properties of these operations is discussed. In \S\ref{distributivity},
the distributivity principles for the $\otimes$ and $\oplus$ families are
generalized to include the four negative operations. In \S\ref{curryhoward},
the (dual) Galois connected operations are given a continuation-passing-style
translation. Linguistic applications of the new vocabulary are discussed
in \S\ref{illustrations}. We conclude with some directions for further 
research.

\section{Residuation and Galois connections}\label{resgal}
Let us recall some key concepts from \cite{dunn:gaggle,residuatedlattices}.
Consider two posets $(X,\leq)$, $(Y,\leq')$ with mappings
$f:X\longrightarrow Y$, $g:Y\longrightarrow X$. The pair $(f,g)$ is called
a residuated pair (\emph{rp}), a dual residuated pair (\emph{drp}), a
Galois connection (\emph{gc}), a dual Galois connection (\emph{dgc}) depending
on which of the following biconditionals holds:
\renewcommand{\arraystretch}{1.2}
\[\begin{array}{rc@{\quad\Leftrightarrow\quad}c}
(\textit{rp}) & fx\leq' y & x\leq gy\\
(\textit{drp}) & y\leq' fx & gy\leq x\\
(\textit{gc}) & y\leq' fx & x\leq gy\\
(\textit{dgc}) & fx\leq' y & gy\leq x\\
\end{array}
\]
Instead of the above biconditionals, one can use an
alternative characterization in terms of the tonicity properties
and the properties of the compositions of the operations involved:
\[\begin{array}{rl@{\quad}l@{\quad}l}
(\textit{rp}) & f,g:\textrm{isotone}, & x\leq gfx, & fgy\leq' y\\
(\textit{drp}) & f,g:\textrm{isotone}, & gfx\leq x, & y\leq' fgx\\
(\textit{gc}) & f,g:\textrm{antitone}, & x\leq gfx, & y\leq' fgy\\
(\textit{dgc}) & f,g:\textrm{antitone}, & fgx\leq x, & gfy\leq' y\\
\end{array}
\]
\renewcommand{\arraystretch}{1}

In the context of categorial type logic, we speak about types and
derivability between types, i.e.~we consider just one inequality.\footnote{Completeness
with respect to relational semantics is discussed in \cite{arecesBM} for the
Galois connected operations, and in \cite{kurtomm07} for the $\otimes/\oplus$
families.}
For the residuated operators of Lambek's syntactic calculus, one
can read $f$ as the operation of multiplying to the right with some
fixed type; $g$ then is right division by that type. The
composition law $fgy\leq' y$ takes the form of the familiar
rightward application schema $(A\slash B)\otimes B\rightarrow A$.
By $\cdot^{\bowtie}$ symmetry, multiplication to the left and
left division similarly form a residuated pair. By arrow
reversal under $\cdot^{\infty}$, we obtain the dual residuated
pairs.

In addition to these binary operations, residuated with respect to
each of their operands, one can also introduce the \emph{unary} case
of residuated pairs in the categorial type language, although neither
Lambek nor Grishin have done so. The defining biconditional is
\[(\textit{rp})\qquad\arr{\fdia A}{B} \quad\Leftrightarrow\quad \arr{A}{\gbox' B}\]
The use of such a pair has been advocated in \cite{morr:type94}
to impose \emph{island constraints} in order to block overgeneration resulting
from the structural rule of associativity. In \cite{kurtonina-moortgat:1997},
the residuated unary operators are used to establish embedding results,
showing that in moving from associative/commutative \textbf{LP} to the
non-associative/non-commutative base logic \textbf{NL}
no expressivity is lost: associativity and/or commutativity can be
recovered in a \emph{controlled} form. On another festive occasion \citep{lambek2007should},
the recipient of this Festschrift has spoken stern words
about the infatuation with diamonds and boxes that one finds in 
certain categorial circles, so I will say no more about them in this paper.

Let us rather turn to monotone \emph{decreasing}
type-forming operations. Such operations are already familiar from the
binary vocabulary, where the (co)implications
among themselves form (dual) Galois connected pairs satisfying
$\arr{A}{C/B}$ iff $\arr{B}{A\bs C}$ and
$\arr{B\obslash C}{A}$ iff $\arr{C\oslash A}{B}$, as we saw.
If the language also contains multiplicative units for $\otimes$
and $\oplus$, one obtains four negations defined in terms of
(co)implication with respect to these units: $\textbf{1}\oslash A$,
$A\bs\textbf{0}$, and the $\bowtie$-symmetric pair. This is the
way the negations are introduced in \cite{grishin83}. A multiplicative
unit for product is not unproblematic for the linguistic applications:
it allows for typing of the empty string or structure which easily
leads to overgeneration. A simple way of avoiding such problems is
to keep the language unit-free and
to introduce the antitone operations as unary connectives in their
own right. For a Galois connected pair $\lneg{\cdot},\rneg{\cdot}$
this was done in \cite{arecesBM}. Here we add a $\infty$-symmetric
dual Galois connected pair $\rdneg{\cdot},\ldneg{\cdot}$. The
Galois principles for these operations
manifest themselves in the following form.
\[(\textit{gc})\quad\arr{B}{\rneg{A}}\,\Leftrightarrow\,\arr{A}{\lneg{B}}\quad ;\quad
(\textit{dgc})\quad\arr{\ldneg{B}}{A}\,\Leftrightarrow\,\arr{\rdneg{A}}{B}\]
The compositions of $\lneg{\cdot},\rneg{\cdot}$ (in either order),
and similarly of $\rdneg{\cdot},\ldneg{\cdot}$, are isotone and idempotent.
For the Galois connected operations, the compositions are expanding;
for the dual Galois operations, they are contracting, i.e.~we have the
arrows below. Together with
monotonicity and idempotence, this means composing the Galois connected
negations yields a \emph{closure} operation; dually, from the composition of
$\rdneg{\cdot},\ldneg{\cdot}$ one obtains an
\emph{interior} operation.
\[\arr{A}{\lneg{(\rneg{A})}}\ ,\ 
\arr{A}{\rneg{(\lneg{A})}}\quad;\quad
\arr{\rdneg{(\ldneg{A})}}{A}\ ,\ 
\arr{\ldneg{(\rdneg{A})}}{A}
\]

\section{Distributivity principles}\label{distributivity}
The properties discussed above depend exclusively on the (dual) Galois principles.
The next natural step is to investigate possible forms of interaction between
the negative operations and the rest of the vocabulary. Our aim here is to
keep the four negations \emph{distinct}, rather than to opt for collapse
into one pair of a cancelling pre- and postnegation \citep{abrusci2002classical,lamb:from93}, or a single
involutive negation \citep{degroote2002cna}.

For communication between the $\otimes$ and $\oplus$ families,
Grishin proposes two groups of interaction principles.
We present them in the rule format of \cite{moot07} and \cite{jfak60}.
One group consists of the rules in Figure \ref{grishindistr}, which
we will collectively refer to as \emph{(distr)}. The other group,
$(\textit{distr})^{-1}$, is
obtained by taking the converses of the inference rules of
Figure \ref{grishindistr}, with premise and conclusion changing place.

\begin{figure}
\begin{center}

\newcommand{\nnd}[2]{#1\rightarrow #2}
\[\begin{array}{c}
\infer[]{\nnd{C\mathbin{\obslash} A}{D\mathbin{\slash} B}}{\nnd{A\otimes B}{C\oplus D}}
\quad\quad
\infer[]{\nnd{B\mathbin{\oslash} D}{A\mathbin{\bs} C}}{\nnd{A\otimes B}{C\oplus D}}\\[2ex]
\infer[]{\nnd{C\mathbin{\obslash} B}{A\mathbin{\bs} D}}{\nnd{A\otimes B}{C\oplus D}}
\quad\quad
\infer[]{\nnd{A\mathbin{\oslash} D}{C\mathbin{\slash} B}}{\nnd{A\otimes B}{C\oplus D}}\\
\end{array}\]

\caption{Distributivity principles \emph{(distr)}}
\label{grishindistr}
\end{center}
\end{figure}

From the principles in \emph{(distr)}, using the (dual) residuation
principles, one easily derives the type transitions below. They change
the dominance relation between the product and the difference operation:
whereas the difference operation is dominated by the product on the
left of the arrow, on the right the difference operation is the main
connective.
\[\begin{array}{l}
 \arr{(A\obslash B)\otimes C}{A\obslash(B\otimes C)}\qquad
\arr{C\otimes(B\oslash A)}{(C\otimes B)\oslash A}\\
 \arr{C\otimes(A\obslash B)}{A\obslash(C\otimes B)}\qquad
\arr{(B\oslash A)\otimes C}{(B\otimes C)\oslash A}\\ 
\end{array}\]
From the $(\textit{distr})^{-1}$ principles, one derives the
type transitions below. For the interaction between product and
difference operations, these are the converses of the above.
\newcommand{\ndr}[2]{#2\rightarrow #1}
\[\begin{array}{l}
 \ndr{(A\obslash B)\otimes C}{A\obslash(B\otimes C)}\qquad
\ndr{C\otimes(B\oslash A)}{(C\otimes B)\oslash A}\\
 \ndr{C\otimes(A\obslash B)}{A\obslash(C\otimes B)}\qquad
\ndr{(B\oslash A)\otimes C}{(B\otimes C)\oslash A}\\ 
\end{array}\]
For interaction between $\otimes$ and $\oplus$, the $(\textit{distr})^{-1}$
principles have the following effect.
\[\begin{array}{l}
 \arr{(A\oplus B)\otimes C}{A\oplus (B\otimes C)}\qquad
\arr{C\otimes(B\oplus A)}{(C\otimes B)\oplus A}\\
 \arr{C\otimes(A\oplus B)}{A\oplus(C\otimes B)}\qquad
\arr{(B\oplus A)\otimes C}{(B\otimes C)\oplus A}\\ 
\end{array}\]

How can we generalize the distributivity principles to include the (dual) Galois
connected operations? In the case where these operations are defined in terms of
multiplicative units this is straightforward: in $(\textit{distr})$ or $(\textit{distr})^{-1}$,
one replaces a subformula of the $\otimes$ term by $\textbf{1}$ and/or of the
$\oplus$ term by $\textbf{0}$. We can extrapolate from the patterns
involving the multiplicative units to obtain versions appropriate
for our unit-free setting. We illustrate with the $(\textit{distr})$ principles.
Compare the derivations below
for a case of interaction among the (dual) Galois connected operators:
\[\infer[(\textit{distr})]{\nd{B\obslash\textbf{1}}{\textbf{0}/A}}{
\infer=[]{\nd{\textbf{1}\otimes A}{B\oplus\textbf{0}}}{\nd{A}{B}}}
\qquad\leadsto\qquad
\infer[]{\nd{\rdneg{B}}{\lneg{A}}}{\nd{A}{B}}\]
Interaction between Galois connected and residuated families
takes the following form:
\[\infer[(\textit{distr})]{\nd{B\obslash\textbf{1}}{C/A}}{
\infer=[]{\nd{\textbf{1}\otimes A}{B\oplus C}}{\nd{A}{B\oplus C}}}
\qquad\leadsto\qquad
\infer[]{\nd{\rdneg{B}}{C/A}}{\nd{A}{B\oplus C}}\]
Taking into account the $\bowtie$ and $\infty$ symmetries,
we obtain the generalized $(\textit{distr})$ principles of
Figure \ref{distrn}.
\begin{figure}
\begin{center}
\[
\infer[]{\arr{\ldneg{B}}{\rneg{A}}}{\arr{A}{B}}\,,\,
\infer[]{\arr{\ldneg{B}}{\lneg{A}}}{\arr{A}{B}}\,,\,
\infer[]{\arr{\rdneg{B}}{\lneg{A}}}{\arr{A}{B}}\,,\,
\infer[]{\arr{\rdneg{B}}{\rneg{A}}}{\arr{A}{B}}
\]
\[
\infer[]{\arr{\rdneg{B}}{A\bs{}C}}{\arr{A}{B\oplus{}C}}\,,\,
\infer[]{\arr{\rdneg{B}}{C\slash{}A}}{\arr{A}{B\oplus{}C}}\,,\,
\infer[]{\arr{\ldneg{C}}{A\bs{}B}}{\arr{A}{B\oplus{}C}}\,,\,
\infer[]{\arr{\ldneg{C}}{B\slash{}A}}{\arr{A}{B\oplus{}C}}
\]
\[
\infer[]{\arr{C\obslash{}A}{\lneg{B}}}{\arr{A\otimes{}B}{C}}\,,\,
\infer[]{\arr{A\oslash{}C}{\lneg{B}}}{\arr{A\otimes{}B}{C}}\,,\,
\infer[]{\arr{C\obslash{}B}{\rneg{A}}}{\arr{A\otimes{}B}{C}}\,,\,
\infer[]{\arr{B\oslash{}C}{\rneg{A}}}{\arr{A\otimes{}B}{C}}
\]

\caption{Generalization of $(\textit{distr})$ for $\cdot^{\textbf{1}},{}^{\textbf{1}}\cdot,\cdot^{\textbf{0}},{}^{\textbf{0}}\cdot$.}
\label{distrn}
\end{center}
\end{figure}

Characteristic theorems depending on the principles of Figure \ref{distrn}
are the laws of the excluded middle below. They follow from the first row of inferences
with the premise instantiated as the identity arrow.
Compare the version with multiplicative units, where
these become $\arr{\textbf{1}\oslash A}{A\bs\textbf{0}}$ iff $\arr{\textbf{1}}{(A\bs\textbf{0})\oplus A}$
(not-$A$ or $A$) etc.
\[\arr{\ldneg{A}}{\rneg{A}}\quad,\quad
\arr{\ldneg{A}}{\lneg{A}}\quad,\quad
\arr{\rdneg{A}}{\lneg{A}}\quad,\quad
\arr{\rdneg{A}}{\rneg{A}}
\]
As long as one makes a choice for either the $(\textit{distr})$ or the
$(\textit{distr})^{-1}$ group of distributivity principles, the four negations
remain distinct operations. Grishin himself follows a different route: to
the mixed-associativity laws of $(\textit{distr})^{-1}$, he adds the
corresponding excluded middle laws as extra axioms, leading to the
identifications $\textbf{1}\oslash A\leftrightarrow A\bs\textbf{0}$
and $\textbf{0}\slash A\leftrightarrow A\obslash \textbf{1}$.
The mixed-associativity laws of $(\textit{distr})$ then become
derivable, i.e.~the distributivity rules become invertible. For the
linguistic applications we have in mind, invertibility of the
distributivity rules is not an option: we need the full group of distributivities
(mixed associativity \emph{and} mixed commutativity laws); invertible 
distributivity rules in that situation mean that the non-associativity/non-commutativity of 
the $\otimes$ and $\oplus$ operations is no longer preserved, as shown in
\cite{bastenhofFG}.

\begin{figure}
\begin{center}
\includegraphics[width=\textwidth]{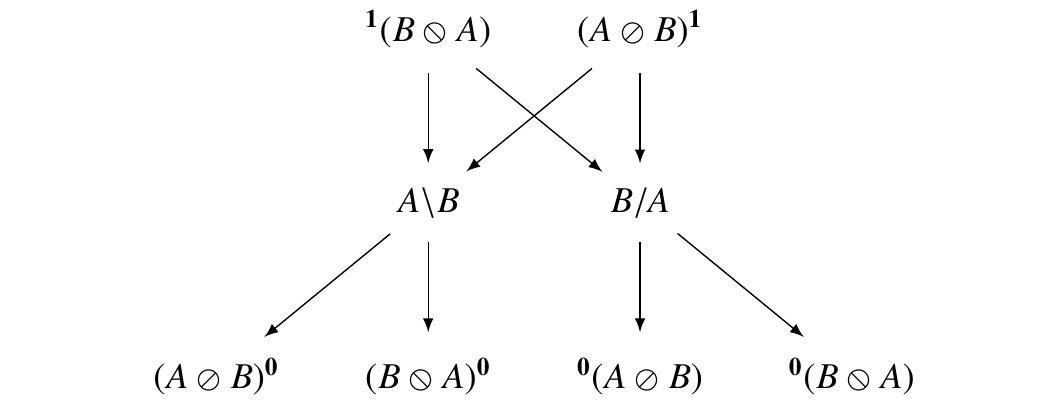}











\caption{Some consequences of the $(\textit{distr})$ principles}
\label{slashco}
\end{center}
\end{figure}

With respect to the de Morgan laws and the expressibility of the (co)implications
in terms of (co)product and negation, the choice between the $(\textit{distr})$ or
$(\textit{distr})^{-1}$ principles again leads to one-way arrows rather than
equalities. For the de Morgan laws, from the $(\textit{distr})$ principles
one derives the inequalities below (and variants with
${\bowtie}$-symmetric formulas on the left and/or on the right of
the arrow).
\renewcommand{\arraystretch}{1}
\[\begin{array}{c}
\arr{\rdneg{(A\otimes B)}}{\lneg{B}\oplus\lneg{A}}\\
\arr{\rdneg{(A\otimes B)}}{\lneg{A}\oplus\lneg{B}}\\
\end{array}
\qquad
\begin{array}{c}
\arr{\rdneg{A}\otimes\rdneg{B}}{\lneg{(B\oplus A)}}\\
\arr{\rdneg{B}\otimes\rdneg{A}}{\lneg{(B\oplus A)}}\\
\end{array}
\]
Inequalities of the following type then express the
relation between (co)implication and (co)product plus negation.
\[\begin{array}{c}
\arr{A\bs B}{\rneg{A}\oplus B}\\
\arr{A\bs B}{B\oplus\rneg{A}}\\
\end{array}
\qquad
\begin{array}{c}
\arr{B\otimes\ldneg{A}}{B\oslash A}\\
\arr{\ldneg{A}\otimes B}{B\oslash A}\\
\end{array}
\]
In Figure \ref{slashco}, we give the neighbours of $A\bs B$ and $B\slash A$
in terms of the (dual) Galois negations, given $(\textit{distr})$. A vertical $\bowtie$ symmetry
axis runs through the middle of the picture. For $B\oslash A$ and $A\obslash B$,
the dual situation obtains: take the $\infty$-symmetric image of the formulas, and
turn around the arrows.

With a choice for $(\textit{distr})^{-1}$, the arrows in the above inequalities
are turned around. We don't elaborate on this option, because the illustrations we'll
discuss in \S\ref{illustrations} only make use of the $(\textit{distr})$ principles.
Before turning to these illustrations, we extend the Curry-Howard interpretation
to the (dual) Galois connected operations.

\renewcommand{\arraystretch}{1}
\section{Proofs and terms}\label{curryhoward}
As argued in \S\ref{intro}, the computational semantics of \LG\ takes
the form of a continuation-passing-style (CPS) translation associating the
derivations of our multiple-conclusion source logic with derivations
of single-conclusion \textbf{LP}. The latter are Curry-Howard isomorphic
with terms of the linear lambda calculus. Our purpose in this section
is to extend the call-by-value CPS translation for (the (co)implication
fragment of) \LG\ of \cite{bernardimm09} to the (dual) Galois negations.
To this end, we present \LG\ in the format of a Display Logic, and we 
define a mapping $\CBV{\cdot}$ acting on its types and derivations:
\[\CBV{\cdot}:
\textbf{LG}^{\mathcal{A}}_{/,\bs,\oslash,\obslash,
\cdot^{\textbf{1}},{}^{\textbf{1}}\cdot,
\cdot^{\textbf{0}},{}^{\textbf{0}}\cdot}
\longrightarrow
\textbf{LP}^{\mathcal{A}\cup\{r\}}_{\Ra}\]

\paragraph{Types} The target calculus has the same atoms as the
source, plus a distinguished atom $r$, the response type. The source
calculus connectives are all interpreted in terms of linear implicative
types with head type $r$. We write $A^{\perp}$ for $A\Ra r$. 
For source types $A$, the target language distinguishes values $\CBV{A}$,
continuations $\CBV{A}^{\perp}$ and computations $\CBV{A}^{\perp\perp}$.
Because the target logic is non-directional, the translation identifies
left-right symmetric source types:
$\CBV{A}=\CBV{A^{\bowtie}}$. For atoms $p\in\mathcal{A}$, $\CBV{p}=p$.
For complex types, we have the mapping below.
\[\CBV{A\bs B}=\CBV{B}^{\perp}\Ra\CBV{A}^{\perp}\ ;\ 
\CBV{A\oslash B}=\CBV{A\bs B}^{\perp}\ ;\ 
\CBV{A^{\textbf{0}}}=\CBV{{}^{\textbf{1}}A}=\CBV{A}^{\perp}\]

\paragraph{Proofs and terms} The presentation of \LG\ as a display sequent
calculus in \cite{jfak60} essentially follows \cite{gore},\footnote{The relation between
display calculus and the Gentzen-style categorial sequent calculi
is discussed in \cite{areces2004analyzing}.} but adds a
mechanism to make a distinction between active and passive formulas. 
Sequent structures are built out of labeled formulas, considered passive:
\emph{input} formulas (hypotheses) are labeled with variables $x$, $y$, $z$, \ldots,
\emph{output} formulas (conclusions) with covariables $\alpha$, $\beta$, $\gamma$, \ldots.
A characteristic feature of the Display Logic format is that for \emph{every} logical
connective (not just for product and coproduct) there is a matching structural connective.
We opt for clarity rather than economy of notation, and use the same symbols for logical and structural
operations, marking off the latter by means of center dots. Input structures $\mathcal{I}$ and
output structures $\mathcal{O}$ are then built according to the grammar below.
  
\[\begin{array}{l@{\quad::=\quad}l}
\mathcal{I} & x:A \mid
	\mathcal{I}\cdot\otimes\cdot\mathcal{I}
	\mid
	\mathcal{I}\cdot\oslash\cdot\mathcal{O}
	\mid
	\mathcal{O}\cdot\obslash\cdot\mathcal{I} \mid 
	{}^{1\cdot}\mathcal{O} \mid \mathcal{O}^{\,\cdot 1}\\
\mathcal{O} & \alpha:A \mid
	\mathcal{O}\cdot\oplus\cdot\mathcal{O}
	\mid
	\mathcal{I}\cdot\bs\cdot\mathcal{O}
	\mid
	\mathcal{O}\cdot\slash\cdot\mathcal{I} \mid 
	\mathcal{I}^{\,\cdot 0} \mid {}^{0\cdot}\mathcal{I} \\
\end{array}
\]
The (dual) residuation and (dual) Galois principles can now be
formulated at the structural level. We don't repeat these rules: simply
replace the formula variables $A$, $B$, \ldots of the arrow presentation
by structure variables $X$, $Y$, \ldots (with input or output interpretation
depending on the context) and the logical connectives by
their structural counterpart. For example,

\[\infer={\arr{A\otimes B}{C}}{\arr{A}{C/B}}
\qquad\leadsto\qquad
\infer={\nd{X\otimesS Y}{Z}}{\nd{X}{Z\slashS Y}}
\]
These rules are invertible; they allow you to display any formula making up
a structure as the single occupant of the sequent antecedent or succedent, depending
on its input/output polarity. Sequents related by the (dual) residuation or Galois rules we
call \emph{display equivalent}.
The distributivity principles, likewise, take the form of structural 
rules in the Display Logic presentation. For example,

\[\infer{\arr{C\obslash A}{D/B}}{\arr{A\otimes B}{C\oplus D}}
\qquad\leadsto\qquad
\infer{\nd{Z\obslashS X}{W\slashS Y}}{\nd{X\otimesS Y}{Z\oplusS W}}
\]

As said, we make a distinction between active and passive
formulas. A sequent can have at most one active formula, which is unlabeled and
displayed as the sole antecedent or succedent formula. In all, then, this gives us
three kinds of sequent: $\nd{X}{Y}$ (all formulas are passive),
$\nd{X}{A}$ (active output formula), $\nd{A}{Y}$ (active input formula).
As will become clear below, there are explicit inference rules to
activate a passive formula, on the input or on the output side.

In \cite{bernardimm09} proofs of the source calculus are coded by
their own term language, a suitably adapted version of the \LaMu\ calculus
of \cite{curi:dual00}.
Here we define the CPS translation directly on the proofs of the
source. The target calculus consists of natural deduction proofs
in correspondence with a fragment of the linear lambda calculus. 
The translation respects the following invariants:

\begin{itemize}
  \item target judgements are of the form $\Gamma\vdash M:B$, where $\Gamma$, the typing environment for
  the target terms, is a multiset of type declarations $\widetilde{x}:\CBV{A}$
(resp.~$\widetilde{\alpha}:\CBV{A}^{\perp}$)
for the passive input (resp.~output) formulas making up the
structures appearing in the source proofs;
  \item source sequents $X\vdash Y$ are mapped to target terms of type $r$; structural
  rules rewriting $X\vdash Y$ to $X'\vdash Y'$ leave the associated term unaffected;
  \item source sequents $X\vdash A$ are mapped to terms of type $\CBV{A}^{\perp\perp}$
(computations);
\item source sequents $A\vdash Y$ are mapped to
terms of type $\CBV{A}^{\perp}$ (continuations).
\end{itemize}

Below we present the rules of the source calculus, followed by their
$\CBV{\cdot}$ translation. First the identity group ((Co)Axiom, Cut) and
the rules for activating a displayed passive formula.

\renewcommand{\arraystretch}{2}
\addtolength{\inferLineSkip}{2pt}

\[\begin{array}{c@{\qquad}c@{\qquad}c}
\infer[\textrm{Ax}]{\nd{x:A}{A}}{} & 
\infer[\textrm{Cut}]{\nd{X}{Y}}{\nd{X}{A}&\nd{A}{Y}} &
\infer[\textrm{Co-Ax}]{\nd{A}{\alpha:A}}{}\\
\end{array}
\]
\[\begin{array}{c@{\qquad}c}
\infer[\comu]{\nd{A}{Y}}{\nd{x:A}{Y}} & \infer[\mu]{\nd{X}{A}}{\nd{X}{\alpha:A}}\\
\end{array}\]

\renewcommand{\arraystretch}{1.5}
\[\begin{array}{c@{\qquad}c}
\CBV{\textrm{Ax}} = \lambda k.(k\ \widetilde{x}):\CBV{A}^{\perp\perp} & 

\CBV{\textrm{Co-Ax}} = \widetilde{\alpha}:\CBV{A}^{\perp}\\
\end{array}
\]
\[\CBV{\textrm{Cut}} = (M^{\CBV{A}^{\perp\perp}} K^{\CBV{A}^{\perp}}):r\]
\[\begin{array}{c@{\qquad}c}
\CBV{\comu} = \lambda\widetilde{x}.S^{\displaystyle r}:\CBV{A}^{\perp} &
\CBV{\mu} = \lambda\widetilde{\alpha}.S^{\displaystyle r}:\CBV{A}^{\perp\perp}
\end{array}
\]

The \emph{logical} rules of the source calculus introduce an active input or
output formula in the conclusion. Rules with a passive premise
simply replace a structural connective by the corresponding logical
one. Rules with active premise(s) compose the active formula
of the conclusion out of the active subformula(e) of the premise(s).

Below the rules for the (dual) Galois negations and their
translations. In the case of $({}^{\textbf{0}}\!\cdot L)$, we can
have the identity transformation, because
$\CBV{A}^{\perp\perp}=\CBV{{}^{\textbf{0}}A}^{\perp}$: the term
coding the premise, a \emph{computation} of type $A$, can also be
interpreted as a \emph{continuation} of type ${}^{\textbf{0}}A$,
as required for the term coding the conclusion.
\renewcommand{\arraystretch}{2}
\[\begin{array}{c@{\qquad}c}
\infer[{\cdot}^{\textbf{1}}L]{\nd{A^{\textbf{1}}}{Y}}{\nd{(\alpha:A)^{\,\cdot 1}}{Y}} &
\infer[{\cdot}^{\textbf{1}}R]{\nd{Y^{\,\cdot 1}}{A^{\textbf{1}}}}{\nd{A}{Y}}\\
\infer[{}^{\textbf{0}}\!\cdot R]{\nd{X}{{}^{\textbf{0}}A}}{\nd{X}{{}^{0\cdot}(x:A)}} & 
\infer[{}^{\textbf{0}}\!\cdot L]{\nd{{}^{\textbf{0}}A}{{}^{0\cdot}X}}{\nd{X}{A}}\\
\end{array}
\]
\renewcommand{\arraystretch}{1.5}
\[\begin{array}{l@{\qquad}l}
\CBV{{\cdot}^{\textbf{1}}L} = \lambda\widetilde{\alpha}.S^{\displaystyle r}:\CBV{A^{\textbf{1}}}^{\perp} &
\CBV{{\cdot}^{\textbf{1}}R} =\lambda k.(k\ K^{\CBV{A}^{\perp}}):\CBV{A^{\textbf{1}}}^{\perp\perp}\\
\CBV{{}^{\textbf{0}}\!\cdot R} = \lambda k.(k\ \lambda\widetilde{x}.S^{\displaystyle r}):\CBV{{}^{\textbf{0}}A}^{\perp\perp} &
\CBV{{}^{\textbf{0}}\!\cdot L} = M^{\CBV{A}^{\perp\perp}}:\CBV{{}^{\textbf{0}}A}^{\perp}\\
\end{array}
\]
Finally, the rules for the (co)implications. We give the 
rules for formulas $A\bs B$ and $A\oslash B$ (rather than $B\oslash A$, which is the dual of $A\bs B$)
in order to highlight the correspondence between the interpretation of implication and co-implication.
\renewcommand{\arraystretch}{2}
\[\begin{array}{cc}
\infer[\bs R]{\nd{X}{A\bs B}}{\nd{X}{(x:A)\bsS(\beta:B)}} &
\infer[\oslash L]{\nd{A\oslash B}{X}}{\nd{(x:A)\oslashS(\beta:B)}{X}} \\[1ex]
\infer[\bs L]{\nd{A\bs B}{X\bsS Y}}{\nd{X}{A} & \nd{B}{Y}} &
\infer[\oslash R]{\nd{X\oslashS Y}{A\oslash B}}{\nd{X}{A} & \nd{B}{Y}} \\
\end{array}
\]
\renewcommand{\arraystretch}{1.5}
\[\begin{array}{r@{\quad=\quad}l}
\CBV{\bs R}=\CBV{\oslash L} & \lambda h.(h\ \lambda\widetilde{\beta}\lambda\widetilde{x}.S^{\displaystyle r}):\CBV{A\bs B}^{\perp\perp}=\CBV{A\oslash B}^{\perp}\\
\CBV{\bs L} & \lambda u.(M^{\CBV{A}^{\perp\perp}} (u\ K^{\CBV{B}^{\perp}})):\CBV{A\bs B}^{\perp}\\
\CBV{\oslash R} & \lambda k.(k\ \CBV{(\bs L)}):\CBV{A\oslash B}^{\perp\perp}\\
\end{array}
\]

For the binary vocabulary, I have shown in \citep{jfak60} that \textbf{LG} enjoys Cut elimination.
Extending this result to the unary negative operations presents no problems. Below the
transformation for a principal cut on ${}^{\textbf{0}}A$ in the source
calculus together with the image (normalization/$\beta$ conversion) 
under the $\CBV{\cdot}$ translation. The remaining cases are
obtained from the $\bowtie$ and $\infty$ symmetries.

\[\begin{array}{c@{\quad\leadsto\quad}c}
\infer[\textrm{Cut}]{\nd{X}{{}^{0\cdot}Y}}{
\infer[{}^{\textbf{0}}\!\cdot R]{\nd{X}{{}^{\textbf{0}}A}}{\nd{X}{{}^{0\cdot}(x:A)}}
&
\infer[{}^{\textbf{0}}\!\cdot L]{\nd{{}^{\textbf{0}}A}{{}^{0\cdot}Y}}{\nd{Y}{A}}}
&
\infer[gc]{\nd{X}{{}^{0\cdot}Y}}{
\infer[\textrm{Cut}]{\nd{Y}{X^{\,\cdot 0}}}{
\nd{Y}{A} &
\infer[\comu]{\nd{A}{X^{\,\cdot 0}}}{
\infer[gc]{\nd{x:A}{X^{\,\cdot 0}}}{\nd{X}{{}^{0\cdot}(x:A)}}}}}\\

(\lambda k.(k\ \lambda\widetilde{x}.S^{\displaystyle r})\ M^{\CBV{A}^{\perp\perp}})
&
(M^{\CBV{A}^{\perp\perp}}\ \lambda\widetilde{x}.S^{\displaystyle r})\\
\end{array}\]

\renewcommand{\arraystretch}{1}
\renewcommand{\CBVlex}[1]{|#1|}
\section{Illustrations}\label{illustrations}
Let us turn to the possible uses of the negative operations
in combination with the rest of the vocabulary. We give examples of
new expressive facilities that rely exclusively on the residuation and Galois
principles, and examples involving also the distributivity principles
$(\textit{distr})$. To accommodate the \emph{lexical}
recipes of a simple extensional Montague-style interpretation,
we compose the \emph{derivational} semantics given by the CPS translation
with a mapping $\CBVlex{\cdot}$.
\[
\textbf{LG}^{\mathcal{A}}_{/,\bs,\oslash,\obslash,
\cdot^{\textbf{1}},{}^{\textbf{1}}\cdot,
\cdot^{\textbf{0}},{}^{\textbf{0}}\cdot}
\xlongrightarrow{\makebox[.3in]{$\CBV{\cdot}$}}
\textbf{LP}^{\mathcal{A}\cup\{r\}}_{\Ra}
\xlongrightarrow{\makebox[.3in]{$\CBVlex{\cdot}$}}
\textbf{IL}^{\{e,t\}}_{\rightarrow}
\]
On the type level, $\CBVlex{\cdot}$ associates the atomic syntactic types in
$\mathcal{A}$ and the response type $r$ with target semantic types built
from the atomic semantic types $e,t$. For atomic syntactic types in
$\mathcal{A}$, $\CBVlex{\cdot}$ coincides with the mapping from syntactic
to semantic types of a direct (non-continuized) interpretation, with
$\CBVlex{\textit{np}}=e$, $\CBVlex{s}=t$, $\CBVlex{n}= e\Ra t$, for example.
For the continuation response type, let us assume $\CBVlex{r}=t$. 
As a result of the identification $\CBVlex{r}=\CBVlex{s}$, the interpretation
of a sentence computation, $\CBVlex{\CBV{s}^{\perp\perp}}$, will be given
by a term of type $(t\Ra t)\Ra t$. If this sentence stands on its own,
i.e.~if there is no bigger context of which it forms a part, we can
evaluate it to a truth-value denoting expression by providing the
trivial continuation --- the identity function of type $t\Ra t$.

On the level of proofs/terms, source \emph{constants} of type $A$ are
associated with closed target terms of type $\CBVlex{A}$. These \emph{lexical}
recipes are not required to be linear. But on \emph{complex} source types and terms, 
$\CBVlex{\cdot}$ acts homomorphically, so that, apart from possible non-linear
contributions of the lexical items, the linearity of the source terms is
reflected in the translation.
\[
\CBVlex{(M\ N)}=(\CBVlex{M}\ \CBVlex{N})\quad;\quad
\CBVlex{\lambda x.M}=\lambda\widetilde{x}.\CBVlex{M}
\]

\paragraph{Scope} Our first example illustrates the use of the
\emph{interior} operation, i.e.~the composition of the
dual Galois connected operations ${}^{\textbf{1}}(\,\cdot\,^{\textbf{1}})$.
This example makes no use of the distributivity postulates. 
Suppose we assign type ${}^{\textbf{1}}(np^{\textbf{1}})$ to quantifier phrases (`everyone',
`some student', \ldots). The type contracts to $\textit{np}$, accounting for the fact that
such phrases syntactically behave as simple noun phrases. In the case where a
sentence contains multiple quantifier phrases, there is a derivational
ambiguity as to the points in the derivation where the $(\cdot^{\textbf{1}} R)$ rules apply.
These choice points lead to the different scope construals for such a sentence.

Below, we give two derivations, using the compact format introduced in \cite{jfak60}: the
display equivalences and the formula (de)activation steps leading from one active formula to the next are
compiled away; for legibility, only the (co)axiom formulas and the
input values of the endsequent are explicitly labeled.

 Tracing the steps in backward chaining fashion, the two derivations have the same
initial moves: the focus is shifted from the goal formula $s$ first to
the subject, then to the direct object; the main connective in each case is
rewritten to its structural counterpart by the $({}^{\textbf{1}}\!\cdot  L)$ rules.
At that point, the derivations diverge. 
In the case of (\dag), $(\cdot^{\textbf{1}}  R)$
introduces the conegation on the direct object $\textit{np}$.

\[ \infer[\leftrightharpoons]{
\underbrace{{}^{\textbf{1}}(np^{\textbf{1}})}_{\textrm{su}}\cdot\otimes\cdot (
\underbrace{(np_{} \bs s_{})/ np_{}}_{\textrm{tv}}\cdot\otimes\cdot 
\underbrace{{}^{\textbf{1}}(np^{\textbf{1}})}_{\textrm{do}}) \vdash s_{}}{
 \infer[{}^{\textbf{1}}\!\cdot  L]{{}^{\textbf{1}}(np^{\textbf{1}}) \vdash s_{}\cdot\slash\cdot ((np_{} \bs s_{})/ np_{}\cdot\otimes\cdot {}^{\textbf{1}}(np^{\textbf{1}}))}{
 \infer[{}^{\textbf{1}}\!\cdot  L]{{}^{\textbf{1}}(np^{\textbf{1}}) \vdash (np_{} \bs s_{})/ np_{}\cdot\bs\cdot ({}^{1\cdot} (np^{\textbf{1}})\cdot\bs\cdot s_{})}{
 \infer[\cdot^{\textbf{1}}  R\quad\textrm{(direct object)}]{\hspace{-.8cm}\dag\quad((np_{} \bs s_{})/ np_{}\cdot\bs\cdot ({}^{1\cdot} (np^{\textbf{1}})\cdot\bs\cdot s_{})){}^{\cdot 1} \vdash np^{\textbf{1}}}{
 \infer[\leftleftharpoons]{np_{} \vdash (np_{} \bs s_{})/ np_{}\cdot\bs\cdot ({}^{1\cdot} (np^{\textbf{1}})\cdot\bs\cdot s_{})}{
 \infer[\slash L]{(np_{} \bs s_{})/ np_{} \vdash ({}^{1\cdot} (np^{\textbf{1}})\cdot\bs\cdot s_{})\cdot\slash\cdot np_{}}{
 \infer[\bs L]{np_{} \bs s_{} \vdash {}^{1\cdot} (np^{\textbf{1}})\cdot\bs\cdot s_{}}{
 \infer[\rightrightharpoons]{{}^{1\cdot} (np^{\textbf{1}}) \vdash np_{}}{
 \infer[\cdot^{\textbf{1}}  R]{np^{\,\cdot 1} \vdash np^{\textbf{1}}}{
np_{} \stackrel{}{\vdash} \beta:np_{} }} & 
s_{} \stackrel{}{\vdash} \alpha:s_{}} & 
y:np_{}\stackrel{}{\vdash} np_{}}} }}}}\]
%

The stepwise construction of the $\CBV{\cdot}$ translation below shows that this
derivation is mapped to an interpretation where the direct object outscopes the subject.

\renewcommand{\arraystretch}{1.2}

\[\begin{array}{rr@{\quad:\quad}l}
\cdot^{\textbf{1}}R & \lambda k.(k\ \widetilde{\beta}) & \CBV{np^{\textbf{1}}}^{\perp\perp}\\
\rightrightharpoons & \lambda\widetilde{\beta}.(\widetilde{\gamma}\ \widetilde{\beta})=\widetilde{\gamma} & \CBV{np^{\textbf{1}}}^{\perp}=\CBV{np}^{\perp\perp}\\
\bs L & \lambda u.(\widetilde{\gamma}\ (u\ \widetilde{\alpha})) & \CBV{np\bs s}^{\perp}\\
\slash L & \lambda u'.(u'\ \lambda u.(\widetilde{\gamma}\ (u\ \widetilde{\alpha}))\ \widetilde{y}) & \CBV{(np\bs s)/np}^{\perp}\\
\leftleftharpoons & \lambda\widetilde{y}.(\mathrm{tv}\ \lambda u.(\widetilde{\gamma}\ (u\ \widetilde{\alpha}))\ \widetilde{y}) & \CBV{np}^{\perp}\\
\cdot^{\textbf{1}}R & \lambda k.(k\ \lambda\widetilde{y}.(\textrm{tv}\ \lambda u.(\widetilde{\gamma}\ (u\ \widetilde{\alpha}))\ \widetilde{y})) & 
\CBV{np^{\textbf{1}}}^{\perp\perp}\\
{}^{\textbf{1}}\!\cdot  L & \lambda\widetilde{\kappa}.(\widetilde{\kappa}\ \lambda\widetilde{y}.(\textrm{tv}\ \lambda u.(\widetilde{\gamma}\ (u\ \widetilde{\alpha}))\ \widetilde{y})) & 
\CBV{{}^{\textbf{1}}(np^{\textbf{1}})}^{\perp}\\
{}^{\textbf{1}}\!\cdot  L & \lambda\widetilde{\gamma}.(\mathrm{do}\ \lambda\widetilde{y}.(\textrm{tv}\ \lambda u.(\widetilde{\gamma}\ (u\ \widetilde{\alpha}))\ \widetilde{y})) 
& \CBV{{}^{\textbf{1}}(np^{\textbf{1}})}^{\perp}\\
\leftrightharpoons & \lambda\widetilde{\alpha}.(\mathrm{do}\ \lambda\widetilde{y}.((\mathrm{tv}\ \lambda u.(\mathrm{su}\ (u\ \widetilde{\alpha})))\ \widetilde{y}))& \CBV{s}^{\perp\perp}\\
\end{array}
\]

\renewcommand{\arraystretch}{1}

Some comments on the steps. The focus shifting rules $\rightrightharpoons$, $\leftleftharpoons$, $\leftrightharpoons$,
are shorthand for a sequence of steps: first the deactivation of the active formula of the premise,
then display equivalences to bring a new formula in focus, and finally a $\mu$ or $\comu$ step activating
that new formula. Deactivation of the premise active formula is achieved 
by means of a cut against a (co)axiom; these cuts introduce the (co)variables $\widetilde{\gamma}$,
$\mathrm{tv}$, and $\mathrm{su}$. The $\mu$ or $\comu$ steps then build a computation or
continuation term for the conclusion by binding a (co)variable of the appropriate type,
$\widetilde{\beta}$, $\widetilde{y}$, $\widetilde{\alpha}$ in the case at hand. The conclusion of the
$({}^{\textbf{1}}\!\cdot  L)$ rules, similarly, is obtained from an implicit cut on a (co)axiom,
introducing the (co)variables $\widetilde{\kappa}$ and $\mathrm{do}$ of type 
$\CBV{np^{\textbf{1}}}^{\perp}$ and $\CBV{{}^{\textbf{1}}(np^{\textbf{1}})}$ respectively.

So far the direct object wide scope interpretation.
The alternative derivation, shown below, proceeds with (\ddag) where we had (\dag) before.
In the case of (\ddag), the $(\cdot^{\textbf{1}} R)$ rule introduces the conegation
on the subject $\textit{np}$.

\[
\infer[\leftrightharpoons]{{}^{\textbf{1}}(np^{\textbf{1}})\cdot\otimes\cdot ((np_{} \bs s_{})/ np_{}\cdot\otimes\cdot {}^{\textbf{1}}(np^{\textbf{1}})) \vdash s_{}}{
 \infer[{}^{\textbf{1}}\!\cdot  L]{{}^{\textbf{1}}(np^{\textbf{1}}) \vdash s_{}\cdot\slash\cdot ((np_{} \bs s_{})/ np_{}\cdot\otimes\cdot {}^{\textbf{1}}(np^{\textbf{1}}))}{
 \infer[{}^{\textbf{1}}\!\cdot  L]{{}^{\textbf{1}}(np^{\textbf{1}}) \vdash (np_{} \bs s_{})/ np_{}\cdot\bs\cdot ({}^{1\cdot} (np^{\textbf{1}})\cdot\bs\cdot s_{})}{
 \infer[\cdot^{\textbf{1}}  R\quad\textrm{(subject)}]{\hspace{-.8cm}\ddag\quad(s_{}\cdot\slash\cdot ((np_{} \bs s_{})/ np_{}\cdot\otimes\cdot {}^{1\cdot} (np^{\textbf{1}}))){}^{\cdot 1} \vdash np^{\textbf{1}}}{
 \infer[\leftleftharpoons]{np_{} \vdash s_{}\cdot\slash\cdot ((np_{} \bs s_{})/ np_{}\cdot\otimes\cdot {}^{1\cdot} (np^{\textbf{1}}))}{
 \infer[\slash L]{(np_{} \bs s_{})/ np_{} \vdash (np_{}\cdot\bs\cdot s_{})\cdot\slash\cdot {}^{1\cdot} (np^{\textbf{1}})}{
 \infer[\bs L]{np_{} \bs s_{} \vdash np_{}\cdot\bs\cdot s_{}}{
x:np_{} \vdash np_{} & s_{} \vdash\alpha:s_{}} &  
\infer[\rightrightharpoons]{{}^{1\cdot} (np^{\textbf{1}}) \vdash np_{}}{\vdots
}}} }
}
}}
\]
We compute the $\CBV{\cdot}$ translation. The abbreviated right branch is mapped to a term
$\widetilde{\kappa}$ of type $\CBV{np^{\textbf{1}}}^{\perp}=\CBV{np}^{\perp\perp}$, which this
time takes the direct object role. This derivation results in an interpretation where the
subject outscopes the direct object.

\renewcommand{\arraystretch}{1.2}

\[\begin{array}{rr@{\quad:\quad}l}
\bs L & \lambda u.(u\ \widetilde{\alpha}\ \widetilde{x}) & \CBV{np\bs s}^{\perp}\\
\slash L & \lambda u'.(\widetilde{\kappa}\ (u'\ \lambda u.(u\ \widetilde{\alpha}\ \widetilde{x}))) & \CBV{(np\bs s)/np}^{\perp}\\
\leftleftharpoons & \lambda\widetilde{x}.(\widetilde{\kappa}\ (\mathrm{tv}\ \lambda u.(u\ \widetilde{\alpha}\ \widetilde{x}))) & \CBV{np}^{\perp}\\
\cdot^{\textbf{1}}R & \lambda k.(k\ \lambda\widetilde{x}.(\widetilde{\kappa}\ (\mathrm{tv}\ \lambda u.(u\ \widetilde{\alpha}\ \widetilde{x})))) & 
\CBV{np^{\textbf{1}}}^{\perp\perp}\\
{}^{\textbf{1}}\!\cdot  L & \lambda\widetilde{\kappa}.(\widetilde{\gamma}\ \lambda\widetilde{x}.(\widetilde{\kappa}\ (\mathrm{tv}\ \lambda u.(u\ \widetilde{\alpha}\ \widetilde{x})))) & 
\CBV{{}^{\textbf{1}}(np^{\textbf{1}})}^{\perp}\\
{}^{\textbf{1}}\!\cdot  L & \lambda\widetilde{\gamma}.(\widetilde{\gamma}\ \lambda\widetilde{x}.(\mathrm{do}\ (\mathrm{tv}\ \lambda u.(u\ \widetilde{\alpha}\ \widetilde{x}))))
& \CBV{{}^{\textbf{1}}(np^{\textbf{1}})}^{\perp}\\
\leftrightharpoons & \lambda\widetilde{\alpha}.(\mathrm{su}\ \lambda\widetilde{x}.(\mathrm{do}\ (\mathrm{tv}\ \lambda u.(u\ \widetilde{\alpha}\  \widetilde{x}))))
& \CBV{s}^{\perp\perp}\\
\end{array}
\]

\renewcommand{\arraystretch}{1}

The table below gives the $\CBVlex{\cdot}$ translation of the constants, for
a sample sentence `everyone likes someone', assuming 
a non-logical target constant `like' of type $e\Ra e\Ra t$,
and the logical constants $\exists,\forall$ (ignoring the person/thing distinction).

\[\begin{array}{ll}
\textrm{source} & \CBVlex{\cdot}\textrm{ translation} \\\hline
\textrm{everyone}:\CBV{np}^{\perp\perp} & \forall:(e\Ra t)\Ra t\\
\textrm{someone}:\CBV{np}^{\perp\perp} & \exists:(e\Ra t)\Ra t\\
\textrm{likes}:(\CBV{s}^{\perp}\Ra\CBV{np}^{\perp})^{\perp}\Ra\CBV{np}^{\perp} & 
\lambda v\lambda y.(v\ \lambda c \lambda x.(c\ ((\textrm{like}\ y)\ x)))\\
 & :(((t\Ra t)\Ra e\Ra t)\Ra t)\Ra e\Ra t\\
\end{array}\] 
The familiar Montague-style interpretations result from the
composition of the $\CBVlex{\cdot}$ and $\CBV{\cdot}$ translations,
and a final evaluation step, providing the identity function $\lambda p.p$
for the abstraction over the parameter $c$ of type $t\Ra t$.

\[\begin{array}{l}
\CBVlex{\lambda\widetilde{\alpha}.(\mathrm{do}\ \lambda\widetilde{y}.((\mathrm{tv}\ \lambda u.(\mathrm{su}\ (u\ \widetilde{\alpha})))\ \widetilde{y}))}\quad=\\
\qquad\qquad\qquad\lambda c.(\exists\ \lambda y.(\forall\ \lambda x.(c\ ((\textrm{like}\ y)\ x))))\\
\CBVlex{\lambda\widetilde{\alpha}.(\mathrm{su}\ \lambda\widetilde{x}.(\mathrm{do}\ (\mathrm{tv}\ \lambda u.((u\ \widetilde{\alpha})\  \widetilde{x})))) }\quad=\\
\qquad\qquad\qquad\lambda c.(\forall\ \lambda x.(\exists\ \lambda y.(c\ ((\textrm{like}\ y)\ x))))\\
\end{array}
\]
Comparing this analysis of scope-taking with the available alternatives,
we notice that the generalized quantifier type $(e\Ra t)\Ra t$ arises
as the $\CBVlex{\cdot}$ image of the syntactic source type \emph{np}, and the
response type of the continuation semantics: there is no \emph{syntactic}
type $s$ involved. This is in contrast with the usual type assignments
to quantifier phrases, such as $s/(np\bs s)$ in the standard Lambek
calculus, or the `wrapping' alternative $(s\!\uparrow\!np)\!\downarrow\!s$
of Morrill and Valentin (this volume), or again the $(s\oslash s)\obslash np$ assignment
of \cite{bernardimm07}.

The identification of the scope domain with 
syntactic type $s$ has been criticized in \cite{dalrymple1997quantifiers}
on the basis of readings where a quantifier phrase takes scope at a non-sentential
level. Examples would be in situ interpretations of quantifier phrases within
nominal modifiers, or as complements of relational noun constructions (`a solution for
every problem', `every picture of a star'). \cite{tls} obtains $n$-internal local
readings by assigning the modifier head or relational noun a lifted type based on the
syntactic category $s$: 
 $(n\bs n)/((s\!\uparrow\!np)\!\downarrow\!s)$ for the preposition `for',
$n/((s\!\uparrow\!np)\!\downarrow\!s)$ for `picture of'; the lexical semantics
for these items is then given in terms of lower-order constants.

With the negative operations, we can create noun phrase internal scope
possibilities without introducing an artificial syntactic $s$ category.
This time, we use the \emph{expanding} composition of Galois connected operations
${}^{\textbf{0}}(\,\cdot\,^{\textbf{0}})$. With a typing
$n/{}^{\textbf{0}}(np^{\textbf{0}})$ for `picture of', the double negation on the
argument produces a lifted semantic type $(e\Ra t)\Ra t$ under the
combined $\CBV{\cdot}$ and $\CBVlex{\cdot}$ translations. Such doubly
negated arguments would be appropriate also for higher-order transitive
verbs (`seeks', `needs': $(np\bs s)/{}^{\textbf{0}}(np^{\textbf{0}})$) allowing for a \emph{de dicto} versus a \emph{de re}
interpretation of the direct object, and for complement-taking verbs
(`claims', `thinks': $(np\bs s)/{}^{\textbf{0}}(s^{\textbf{0}})$) where both the main clause and the embedded clause
need their own $s$ continuation.

In the table below,
we give the CPS translation of these syntactic source types,
together with their image under $\CBVlex{\cdot}$ and terms
expressing lexical semantics.
At the target end, `pic' is a non-logical constant of type
$e\Ra e\Ra t$.
The target non-logical constants
`seek' and `claim' are of type $((e\Ra t)\Ra t)\Ra e\Ra t$
and $((t\Ra t)\Ra t)\Ra e\Ra t$ respectively. Note that the
$\CBVlex{\cdot}\circ\CBV{\cdot}$ image of $np/n$ for the determiners is
of the appropriate semantic type for the
standard Montagovian lexical recipes.
\[\begin{array}{l@{\ }l}
\textrm{source} & \CBVlex{\cdot}\textrm{ translation} \\\hline
\textrm{picture of} & \lambda k\lambda q.(k\ \lambda x.(q\ \lambda y.(\textrm{pic}\ y\ x) ))\\
\CBV{n}^{\perp}\Ra\CBV{np}^{\perp\perp\perp} & ((e\Ra t)\Ra t)\Ra ((e\Ra t)\Ra t) \Ra t\\[1ex]

\textrm{seeks} & \lambda v\lambda q.(v\ \lambda c \lambda x.(c\ (\textrm{seek}\ q\ x)))\\
(\CBV{s}^{\perp}\!\Ra\!\CBV{np}^{\perp})^{\perp}\!\Ra\!\CBV{np}^{\perp\perp\perp} & (((t\Ra t)\Ra e\Ra t)\Ra t)\Ra ((e\Ra t)\Ra t)\Ra t\\[1ex]
\textrm{claims} & \lambda v\lambda q.(v\ \lambda c \lambda x.(c\ (\textrm{claim}\ q\ x)))\\
(\CBV{s}^{\perp}\!\Ra\!\CBV{np}^{\perp})^{\perp}\!\Ra\!\CBV{s}^{\perp\perp\perp} & (((t\Ra t)\Ra e\Ra t)\Ra t)\Ra ((t\Ra t)\Ra t)\Ra t\\[1ex]
 \textrm{every} & 
 \lambda Q\lambda P.(\forall\ \lambda x.((P\ x)\Rightarrow (Q\ x)))\\
 \CBV{np}^{\perp}\Ra\CBV{n}^{\perp} & (e\Ra t)\Ra(e\Ra t)\Ra t\\[1ex]
 \textrm{some} & 
 \lambda Q\lambda P.(\exists\ \lambda x.((P\ x)\wedge(Q\ x)))\\
  \CBV{np}^{\perp}\Ra\CBV{n}^{\perp} & (e\Ra t)\Ra(e\Ra t)\Ra t\\

\end{array}\]

The lexical entries are put to work to compute some scope ambiguities below.
We give the CPS translation of the derivations, and the result of the $\CBVlex{\cdot}$ 
translation of the constants. We emphasize again that the interpretations we have
discussed so far are obtained on the basis of the \emph{pure} logic of
residuated and Galois connected operations: they do not rely on interaction
principles.

\renewcommand{\textsf}[1]{\textrm{#1}}

\[\begin{array}{ll}
\multicolumn{2}{l}{\textrm{every picture of some teacher}\vdash \textit{np}}\\[1ex]
 & \lambda \widetilde{\alpha}_{}.((\CBVlex{\W{\textsf{pictureof}}} \  (\CBVlex{\W{\textsf{every}}} \  \widetilde{\alpha}_{})) \  \lambda \widetilde{\kappa}_{}.((\CBVlex{\W{\textsf{some}}} \  \widetilde{\kappa}_{}) \  \CBVlex{\W{\textsf{teacher}}}))\\
 
& = \lambda \widetilde{\alpha}_{}.(\textsf{$\forall$} \  \lambda x_{}.(  
\textsf{$\exists$} \  \lambda y_{}.((\textsf{teacher} \  y_{}) \wedge  (\textsf{pic} \  y_{} \  x_{}))
\Rightarrow (\widetilde{\alpha}_{} \  x_{})))\\[1ex]

& \lambda \widetilde{\alpha}_{}.((\CBVlex{\W{\textsf{some}}} \  \lambda \widetilde{y}_{}.((\CBVlex{\W{\textsf{pictureof}}} \  (\CBVlex{\W{\textsf{every}}} \  \widetilde{\alpha}_{})) \  \lambda k_{}.(k_{} \  \widetilde{y}_{}))) \  \CBVlex{\W{\textsf{teacher}}})\\

& =\lambda \widetilde{\alpha}_{}.(\textsf{$\exists$} \  \lambda y_{}.((\textsf{teacher} \  y_{}) \wedge  (\textsf{$\forall$} \  \lambda x_{}.( (\textsf{pic} \  y_{} \  x_{}) \Rightarrow  (\widetilde{\alpha}_{} \  x_{})))))\\
\end{array}
\]
\[\begin{array}{ll}
\multicolumn{2}{l}{\textrm{Alice claims some unicorn left}\vdash s}\\[1ex]
 & \lambda \widetilde{\alpha}_{}.((\CBVlex{\W{\textsf{claims}}} \  \lambda k_{}.(k_{} \  \widetilde{\alpha}_{} \  \CBVlex{\W{\textsf{a.}}})) \  \lambda \widetilde{\kappa}_{}.((\CBVlex{\W{\textsf{some}}} \  (\CBVlex{\W{\textsf{left}}} \  \widetilde{\kappa}_{})) \  \CBVlex{\W{\textsf{uni.}}}))\\
 
 & = \lambda c.(c \  ((\textsf{claims} \  \lambda c'.(\textsf{$\exists$} \  \lambda x_{}.((\textsf{unicorn} \  x_{}) \wedge  (c' \  (\textsf{left} \  x_{})))) \  \textsf{alice})))\\[1ex]
 
 & \lambda \widetilde{\alpha}_{}.((\CBVlex{\W{\textsf{some}}} \  \lambda \widetilde{y}_{}.((\CBVlex{\W{\textsf{claims}}} \  \lambda k_{}.(k_{} \  \widetilde{\alpha}_{} \  \CBVlex{\W{\textsf{a.}}})) \  \lambda \widetilde{\kappa}_{}.(\CBVlex{\W{\textsf{left}}} \  \widetilde{\kappa}_{} \  \widetilde{y}_{}))) \  \CBVlex{\W{\textsf{uni.}}})\\
 
 & = \lambda c.(\textsf{$\exists$} \  \lambda x_{}.((\textsf{unicorn} \  x_{}) \wedge (c \  ((\textsf{claims} \  \lambda c'.(c' \  (\textsf{left} \  x_{}))) \  \textsf{alice}))))\\
 \end{array}
 \]

\paragraph{Infixation} 
Let us turn now to some examples where the distributivity principles do come into play.
For the relation between the binary implication and coimplication
the crucial observation is that from the same premises $\nd{X}{B}$ and $\nd{C}{Y}$,
we can derive an input implication $B\bs C$ or an output coimplication
$B\oslash C$; compare
\[\nd{X\otimesS B\bs C}{Y} \qquad\textrm{versus}\qquad \nd{X}{B\oslash C\oplusS Y}\]
Semantically, we have seen that implication and coimplication combine the same pieces of
information: the latter is interpreted as $\lambda k.(k\ M^{\CBV{B\bs C}})$,
i.e.~the lifted form of the interpretation of the former.   From
a syntactic point of view, there is a difference. The implication $B\bs C$ must
concatenate externally with its argument $X$. But in the case where $X$
is a product structure, the conditions for the application of the $(\textit{distr})$ 
interaction principles are met, and the coimplication can \emph{infix}
itself within $X$ and associate with any of its leafs $A$ into a
formula $(B\oslash C)\obslash A$.

In \cite{bernardimm07} we have shown that this property of nested coimplications
allows us to syntactically model the type schema for in situ binding $q(A,B,C)$ from \cite{moor:gene91}
with a type $(B\oslash C)\obslash A$ (the type $A\oslash (C\obslash B)$ would do as well). An expression with such a type behaves locally as
an $A$ within a domain of type $B$ which is mapped into $C$.
See the derivation below for a `compiled' sequent rule ($qL$). The
notation $X[Y]$ for an input structure singles out a substructure $Y$ of $X$
reachable via a path of structural products. For output structures, we write $X[Y]$
to pick out a substructure $Y$ reachable along a path of structural implications.
With $\widetilde{Y}[\ ]$ we mean the image of the input product context 
$Y[\ ]$ under the residuation inferences. 

\newcommand{\toForm}[1]{#1}

\[\infer[rp]{\nd{\toForm{ Y[\  X[\ (B\oslash C)\obslash A\ ]]}}{D}}{
\infer[\obslash L]{\nd{\toForm{ X[\ (B\oslash C)\obslash A\ ]}}{\widetilde{ Y}[D]}}{
\infer[\textit{distr}^*]{\nd{\toForm{ X[\ (B\oslash C)\cdot\obslash\cdot A\ ]}}{\widetilde{ Y}[D]}}{
\infer[\textit{drp}]{\nd{\toForm{ X[A]}}{(B\oslash C)\cdot\oplus\cdot\widetilde{ Y}[D]}}{
\infer[\oslash R]{\nd{\toForm{ X[A]}\cdot\oslash\cdot\widetilde{ Y}[D]}{B\oslash C}}{
\nd{\toForm{ X[A]}}{B} & \quad 
\infer[rp]{\nd{C}{\widetilde{ Y}[D]}}{
\nd{\toForm{ Y[C]}}{D}}}}}}}
\hspace{-5pt}\leadsto\quad
\infer[qL]{\seq{ Y[\  X[\ q(A,B,C)\ ]]}{D}}{\seq{ X[A]}{B} & \seq{ Y[C]}{D}}
\]

Semantically, there is a difference as to how the types $q(A,B,C)$ and $(B\oslash C)\obslash A$
package the meaning contributions of the subformulae $A$, $B$ and $C$.
Under the direct interpretation, $q(A,B,C)'$, the semantic type
corresponding to $q(A,B,C)$, is $(A'\Ra B')\Ra C'$. Contrast this with the 
CPS interpretation
for $(B\oslash C)\obslash A$, 
\[\CBV{(B\oslash C)\obslash A}\quad=\quad(\CBV{B\bs C}^{\perp\perp}\Ra\CBV{A}^{\perp})^{\perp}\]
which consists essentially of a pair of an $A$ value and a lifted
$B\bs C$ value (i.e.~a $B\oslash C$ continuation). Because our target language
is restricted to the simply typed linear lambda calculus, the pair is
expressed as a curried higher-order function.

Given this CPS interpretation, the $\CBVlex{\cdot}$ translation of an expression
of type $(B\oslash C)\obslash A$ can have the schematic form below
\[\lambda h.((h\ \lambda u.(u\ \CBVlex{M^{\CBV{B\bs C}}}))\ \CBVlex{N^{\CBV{A}}})\]
with $\CBVlex{M}$ and $\CBVlex{N}$ the lexical contributions of the 
$B\bs C$ value and $A$ value respectively. We illustrate with an example
from inflectional morphology. Take a past tense transitive verb `tease+ed'.
Suppose we see the tense morpheme as a function taking a subjectless, non-tensed form
of the verb (type $i$, with interpretation $\CBVlex{i}=e\Ra t$) to a tensed verb phrase with external subject argument
(type $np\bs s$). Tense combines as an affix with the tenseless verbal head,
allowing it to combine with whatever internal arguments (and modifiers) it
may have. For transitive `tease+ed', the lexicon then will contain the following information,
assuming at the target side constants `tease' and `past' of type $e\Ra e\Ra t$ and $t\Ra t$
respectively.
\[\begin{array}{rcl}
\textrm{tease+ed} & :& (i\slash np)\oslash ((np\backslash s)\obslash i)\\[1ex]
\CBVlex{\textrm{tease+ed}} & = &
\lambda h.((h\ \lambda u.(u\ \CBVlex{\textrm{-ed}^{\CBV{(np\backslash s)\slash i}}}))\ \CBVlex{\textrm{tease}^{\CBV{i\slash np}}})\\[1ex]
\CBVlex{\textrm{tease}^{\CBV{i\slash np}}} & = & \lambda Q\lambda y.((Q\ (\textrm{tease}^{e\Ra e\Ra t}\ y))\\[1ex]
\CBVlex{\textrm{-ed}^{\CBV{(np\backslash s)\slash i}}} & = & \lambda V\lambda P.(V\ \lambda c\lambda x.(c\ (\textrm{past}^{t\Ra t}\ (P\ x))))\\
\end{array}
\]
A derivation for `Molly teased Leopold' is given below together with its $\CBV{\cdot}$ 
and $\CBVlex{\cdot}$ translations.
\[ \infer[\leftrightharpoons]{
\underbrace{np}_{\textrm{su}}\cdot \otimes \cdot\,\, (\,\,
\underbrace{(i\slash np)\oslash ((np\backslash s)\obslash i)}_{\textrm{verb+tense}}\,\,\cdot \otimes \cdot 
\underbrace{np}_{\textrm{do}})
\vdash s_{}}{
 \infer[\oslash L]{
 (i\slash np)\oslash ((np\backslash s)\obslash i)\vdash (np\cdot \backslash \cdot s)\cdot \slash \cdot np
 }{
 \infer[\obslash R]{
 (np\cdot \backslash \cdot s)\cdot \obslash \cdot ((i\slash np)\cdot \otimes \cdot np)\vdash (np\backslash s)\obslash i
 }{
 \infer[\leftrightharpoons]{i\slash np_{}\cdot\otimes\cdot np_{} \vdash i_{}}{
 \infer[\slash L]{i_{} \slash np_{} \vdash i_{}\cdot\slash\cdot np_{}}{
\mathop{\cdot}\,np_{}\mathop{\cdot} \stackrel{}{\vdash} np_{}
 & i_{} \stackrel{}{\vdash} \mathop{\cdot}\,i_{}\mathop{\cdot}}} &  \infer[\bs L]{np_{} \bs s_{} \vdash np_{}\cdot\bs\cdot s_{}}{
\mathop{\cdot}\,np_{}\mathop{\cdot} \stackrel{}{\vdash} np_{}
 & s_{} \stackrel{}{\vdash} \mathop{\cdot}\,s_{}\mathop{\cdot}}}}}\]
\[\lambda \widetilde{\alpha}_{}.(\CBVlex{\W{\textrm{verb+tense}}} \  \lambda \widetilde{\beta}_{}.(\lambda \widetilde{z}_{}.(\widetilde{\beta}_{} \  \lambda h_{}.((\widetilde{z}_{} \  (h_{} \  \lambda u_{}.((u_{} \  \widetilde{\alpha}_{}) \  \CBVlex{\W{\textrm{su}}}))) \  \CBVlex{\W{\textrm{do}}}))))\]
\[=\quad\lambda c_{}.(c_{} \  (\textrm{past} \  ((\textrm{tease} \  \textrm{leopold}) \  \textrm{molly})))\]

A type assignment of the form $(B\oslash C)\obslash A$ is appropriate for an infix
functor that associates with a particular host $A$, as in the verb+tense combination. We can use the composition
${}^{\textbf{1}}(\,\cdot\,^{\textbf{0}})$ for infixes that have no such host
requirements, and can be placed freely within their domain of application.
Examples that come to mind are parenthetical adverbs. A lexical type assignment
$s/s$ to an adverb such as `hopefully' only allows it to occur in sentence-initial
position, as in `Hopefully, John left'. With a doubly-negated type assignment 
${}^{\textbf{1}}((s/s)^{\textbf{0}})$, the sentence-initial
position is still available, because of the contraction
${}^{\textbf{1}}((s/s)^{\textbf{0}}) \vdash s/s$, but in addition
the word can occupy any sentence-internal position, as in `John, hopefully, left',
`John left, hopefully'. In the
table below, one finds the continuized interpretation for `adv' with the simple
$s/s$ assignment, using a non-logical constant `hpfy' of type $t\Ra t$ at the
target side, and for `$\textrm{adv}'$' with the doubly-negated type
${}^{\textbf{1}}((s/s)^{\textbf{0}})$. The interpretation for the latter is
simply the lifted form of the interpretation of the former.

\[\begin{array}{ll}
\textrm{source} & \CBVlex{\cdot}\textrm{ translation} \\\hline
\textrm{adv}:\CBV{s}^{\perp}\Ra\CBV{s}^{\perp} & 
\lambda c\lambda p.(c\ (\textrm{hpfy}\ p)):(t\Ra t)\Ra t\Ra t\\
\textrm{adv}':(\CBV{s}^{\perp}\Ra\CBV{s}^{\perp})^{\perp\perp} & 
\lambda k.(k\ \CBVlex{\textrm{adv}}):(((t\Ra t)\Ra t\Ra t)\Ra t)\Ra t\\
\end{array}\]

These examples must suffice to give the reader an idea of the
possible uses of the (dual) Galois connected operations in syntax
and semantics.

\section{Conclusions, further directions}

Where do we go from here?
In this paper we have looked at (dual) Galois connected \emph{unary} type-forming operations. 
As with the (dual) residuated (co)product family, the concept
of Galois connected families generalizes to operations
of greater arity. Below, using \emph{ad hoc} notation, the binary case, with 
a Galois connected triple $\notslash,\boxtimes,\notbackslash$,
and a dual Galois connected triple $\boxslash,\boxplus,\boxbslash$.
The (dual) residuated triples are added for comparison:
mind the direction of the arrows!
The new connectives are downward monotonic in \emph{all} positions.
So far, no linguistic applications have been proposed.

\renewcommand{\arraystretch}{1}

\[\begin{array}{rc@{\quad\Leftrightarrow\quad}c@{\quad\Leftrightarrow\quad}c}
(\textit{rp}) & A\rightarrow C/B & A\otimes B\rightarrow C & B\rightarrow A\bs C\\
(\textit{drp}) & A\leftarrow B\obslash C & B\oplus A\leftarrow C & B\leftarrow C\oslash A\\
(\textit{gc}) & A\rightarrow C\notslash B &A\boxtimes B\leftarrow C & B\rightarrow A\notbackslash C\\
(\textit{dgc}) & A\leftarrow C\boxslash B & A\boxplus B\rightarrow C & B\leftarrow A\boxbslash C\\
\end{array}
\]

A second theme for further research concerns the distributivity principles
$(\textit{distr})$ and $(\textit{distr})^{-1}$. The analysis of \emph{infixation}
phenomena in this paper relies on the $(\textit{distr})$ interactions. 
In \cite{bastenhofFG}, however, one finds an analysis of relativization
on the basis of a type assignment $(n\bs n)/(s\oplus{}^{\textbf{0}}np)$
to the relative pronoun. For \emph{extraction} of the gap, this analysis
uses the $(\textit{distr})^{-1}$ interactions between $\oplus$ and $\otimes$;
these are combined with the $(\textit{distr})$ principles of Fig \ref{distrn}
for the interaction between ${}^{\textbf{0}}$ and $\obslash$. As we saw
above, the $(\textit{distr})$ and $(\textit{distr})^{-1}$ principles cannot
be combined in their full generality without spoiling the non-associative
and non-commutative character of $\otimes/\oplus$. The mixture of
\cite{bastenhofFG} is one way of avoiding overgeneration. The general
picture of a controlled combination of the $(\textit{distr})$ and
$(\textit{distr})^{-1}$ principles is a topic for further research.

\paragraph*{Acknowledgements} My interest in the four unary negation
operations was raised by a comment Jim Lambek
made on an earlier presentation of the symmetric calculus.
Jim's suggestion was to treat the slashes and their duals not
as primitives but as operators \emph{defined} in terms of the
negations and (co)product. For reasons explained above, I split
the defining equations in two symmetric halves so as to have
interaction without loss of structural discrimination.
For comments and discussion, I thank Chris Barker,
Raffaella Bernardi, Arno Bastenhof and Jeroen Bransen.
All errors are my own.

\bibliography{la}

\begin{thebibliography}{25}
\providecommand{\natexlab}[1]{#1}

\bibitem[{Abrusci(2002)}]{abrusci2002classical}
Abrusci, V.M. 2002.
\newblock {Classical conservative extensions of Lambek calculus}.
\newblock \emph{Studia Logica} 71(3): 277--314.

\bibitem[{Areces and Bernardi(2004)}]{areces2004analyzing}
Areces, C. and R.~Bernardi. 2004.
\newblock {Analyzing the core of categorial grammar}.
\newblock \emph{Journal of Logic, Language and Information} 13(2): 121--137.

\bibitem[{Areces \emph{et~al.}(2004)Areces, Bernardi, and Moortgat}]{arecesBM}
Areces, Carlos, Raffaella Bernardi, and Michael Moortgat. 2004.
\newblock Galois Connections in Categorial Type Logic.
\newblock \emph{Electronic Notes in Theoretical Computer Science} 53: 3--20.

\bibitem[{Barker and Shan(2006)}]{barker2006types}
Barker, C. and C.~Shan. 2006.
\newblock {Types as graphs: Continuations in type logical grammar}.
\newblock \emph{Journal of Logic, Language and Information} 15(4): 331--370.

\bibitem[{Bastenhof(2010)}]{bastenhofFG}
Bastenhof, Arno. 2010.
\newblock Polarized {M}ontagovian semantics for the {L}ambek-{G}rishin
  calculus.
\newblock In \emph{Proceedings 15th Conference on Formal Grammar}. Copenhagen.

\bibitem[{Bernardi and Moortgat(2007)}]{bernardimm07}
Bernardi, Raffaella and Michael Moortgat. 2007.
\newblock Continuation semantics for symmetric categorial grammar.
\newblock In \emph{Proceedings 14th Workshop on Logic, Language, Information
  and Computation (WoLLIC'07)}, edited by Daniel Leivant and Ruy de~Queiros,
  \emph{LNCS}, vol. 4576, 53--71. Heidelberg: Springer.

\bibitem[{Bernardi and Moortgat(2010)}]{bernardimm09}
---. 2010.
\newblock Continuation semantics for the {L}ambek-{G}rishin calculus.
\newblock \emph{Information and Computation} 208(5): 397--416.

\bibitem[{Carpenter(1997)}]{tls}
Carpenter, Bob. 1997.
\newblock \emph{{Type-logical semantics}}.
\newblock Cambridge, MA: The MIT Press.

\bibitem[{Curien and Herbelin(2000)}]{curi:dual00}
Curien, P. and H.~Herbelin. 2000.
\newblock Duality of computation.
\newblock In \emph{International Conference on Functional Programming
  (ICFP'00)}, 233--243.
\newblock [2005: corrected version].

\bibitem[{Dalrymple \emph{et~al.}(1997)Dalrymple, Lamping, Pereira, and
  Saraswat}]{dalrymple1997quantifiers}
Dalrymple, M., J.~Lamping, F.~Pereira, and V.~Saraswat. 1997.
\newblock {Quantifiers, anaphora, and intensionality}.
\newblock \emph{Journal of Logic, Language and Information} 6(3): 219--273.

\bibitem[{De~Groote(2001)}]{degr:type01}
De~Groote, P. 2001.
\newblock Type raising, continuations, and classical logic.
\newblock In \emph{Proceedings of the Thirteenth Amsterdam Colloquium}, edited
  by M.~Stokhof R.~van Rooy, 97--101. ILLC, Universiteit van Amsterdam.

\bibitem[{Dunn(1991)}]{dunn:gaggle}
Dunn, J~Michael. 1991.
\newblock Gaggle Theory: An Abstraction of Galois Connections and Residuation
  with Applications to Negation and Various Logical Operators.
\newblock In \emph{Logics in AI, Proceedings JELIA '90 Amsterdam}, edited by
  Jan van Eijck, \emph{Lecture Notes in AI}, vol. 478, 31--51. Heidelberg:
  Springer.

\bibitem[{Galatos \emph{et~al.}(2007)Galatos, Jipsen, Kowalski, and
  Ono}]{residuatedlattices}
Galatos, Nikolaos, Peter Jipsen, Tomasz Kowalski, and Hiroakira Ono. 2007.
\newblock \emph{Residuated Lattices: An Algebraic Glimpse at Substructural
  Logics, Volume 151 (Studies in Logic and the Foundations of Mathematics)}.
\newblock Amsterdam: Elsevier.

\bibitem[{Gor\'e(1997)}]{gore}
Gor\'e, Rajeev. 1997.
\newblock Substructural Logics on Display.
\newblock \emph{Logic Journal of IGPL} 6(3): 451--504.

\bibitem[{Grishin(1983)}]{grishin83}
Grishin, V.N. 1983.
\newblock On a generalization of the {A}jdukiewicz-{L}ambek system.
\newblock In \emph{Studies in Nonclassical Logics and Formal Systems}, edited
  by A.I. Mikhailov, 315--334. Moscow: Nauka.
\newblock [English translation in Abrusci and Casadio (eds.) New Perspectives
  in Logic and Formal Linguistics. Bulzoni, Rome, 2002].

\bibitem[{de~Groote and Lamarche(2002)}]{degroote2002cna}
de~Groote, Philippe and F.~Lamarche. 2002.
\newblock {Classical non-associative {L}ambek calculus}.
\newblock \emph{Studia Logica} 71(3): 355--388.

\bibitem[{Kurtonina and Moortgat(1997)}]{kurtonina-moortgat:1997}
Kurtonina, Natasha and Michael Moortgat. 1997.
\newblock Structural Control.
\newblock In \emph{Specifying Syntactic Structures}, edited by Patrick
  Blackburn and Maarten de~Rijke, 75--113. Stanford: CSLI Publications.

\bibitem[{Kurtonina and Moortgat(2010)}]{kurtomm07}
---. 2010.
\newblock Relational semantics for the {L}ambek-{G}rishin calculus.
\newblock In \emph{MOL 10/11. Selected papers from the 10th and 11th
  Mathematics of Language Meetings, Los Angeles 2007, Bielefeld 2009}, edited
  by Christian Ebert, Gerhard J{\"a}ger, and Jens Michaelis, \emph{LNCS}, vol.
  6149, 210--222. Heidelberg: Springer.

\bibitem[{Lambek(1993)}]{lamb:from93}
Lambek, J. 1993.
\newblock From categorial to bilinear logic.
\newblock In \emph{Substructural Logics}, edited by {Kosta Do\v{s}en and Peter
  Schr\"{o}der-Heister}, 207--237. Oxford University Press.

\bibitem[{Lambek(2007)}]{lambek2007should}
---. 2007.
\newblock {Should Pregroup Grammars be Adorned with Additional Operations?}
\newblock \emph{Studia Logica} 87(2): 343--358.

\bibitem[{Moortgat(1996)}]{moor:gene91}
Moortgat, Michael. 1996.
\newblock Generalized quantifiers and discontinuous type constructors.
\newblock In \emph{Discontinuous Constituency}, edited by Harry Bunt and Arthur
  van Horck, 181--207. Berlin, New York: Mouton De Gruyter.

\bibitem[{Moortgat(2009)}]{jfak60}
---. 2009.
\newblock Symmetric categorial grammar.
\newblock \emph{Journal of Philosophical Logic} 38(6): 681--710.

\bibitem[{Moot(2007)}]{moot07}
Moot, Richard. 2007.
\newblock Proof nets for display logic.
\newblock \emph{CoRR} abs/0711.2444.

\bibitem[{Morrill(1994)}]{morr:type94}
Morrill, Glyn. 1994.
\newblock \emph{Type Logical Grammar}.
\newblock Dordrecht: Kluwer.

\bibitem[{Morrill and Valentin(2010)}]{morrillvalentin}
Morrill, Glyn and Oriol Valentin. 2010.
\newblock Displacement calculus.
\newblock \emph{Linguistic Analysis (this volume)} 36(1--4).

\end{thebibliography}
\nocite{morrillvalentin}

\end{document}